\documentclass[conference]{IEEEtran}
\IEEEoverridecommandlockouts

\usepackage{cite}
\usepackage{amsmath,amssymb,amsfonts}
\usepackage{algorithmic}
\usepackage{graphicx}
\usepackage{textcomp}
\usepackage{xcolor}
\def\BibTeX{{\rm B\kern-.05em{\sc i\kern-.025em b}\kern-.08em
    T\kern-.1667em\lower.7ex\hbox{E}\kern-.125emX}}

\usepackage[utf8]{inputenc}
\usepackage{url}
\usepackage{array}
\usepackage{booktabs}
\usepackage{colortbl}
\usepackage{tabularx}
\usepackage{ragged2e}
\newcolumntype{Y}{>{\RaggedRight\arraybackslash}X}

\begin{document}

\title{Semantic Map Sharing and Capability-Aware Coverage Planning for AI-Native 6G Robotic Coordination\\
\thanks{This research was funded in part by the UKRI Horizon Europe Guarantee Fund (Grant Nos. 10064520 \& 10115919) for the EU Horizon Europe projects P2CODE (101093069) and 6G-PATH (101139172), by the European Commission for the Agentic6G project (101290342) and by UKRI EPSRC for the National Edge AI Hub project (EP/Y028813/1).}
\thanks{$^{1}$School of Computing and Mathematical Sciences, University of Leicester, UK, \{aamd2, d.hao, qw96\}@leicester.ac.uk}
\thanks{$^{2}$A. Dhafer and Z. D. Hao are with DANiLab, University of Leicester, Leicester, UK (corresponding author: d.hao@leicester.ac.uk)}
\thanks{A version of this work was accepted for poster presentation at the 2026 IEEE Conference on Standards for Communications and Networking (IEEE CSCN 2026).}
}

\author{Abdulqader Dhafer$^{1,2}$, Qi Wang$^{1}$, and Zhou Daniel Hao$^{1,2}$}

\maketitle

\begin{abstract}
Search and Rescue (SAR) operations increasingly deploy heterogeneous teams of aerial and ground robots. However, conventional coverage methods typically do not translate perceived terrain into platform-specific reachability, while continuous image exchange imposes a high communication cost. We propose an edge-centric, semantic-aware coverage planning framework that integrates aerial terrain perception, robot-specific traversability reasoning, and payload-efficient semantic state sharing. Aerial observations are converted into compact semantic grid maps, enabling reachability-constrained area decomposition and capability-aware coverage paths that assign only regions admitted by each robot's capability profile. The resulting perception--sharing--planning loop feeds semantic corrections into traversability reasoning and replanning, forming an application-level mechanism motivated by AI-enabled goal-oriented communication envisioned for AI-native 6G networks. For the high-update case, transmitting semantic corrections reduces the application payload by a factor of approximately $82$ relative to periodic full-map sharing. Across matched benchmark scenarios, the proposed method achieved $91.5\%$ coverage with no capability-infeasible allocations, compared with $78.8\%$ coverage and a $21.5\%$ capability-infeasible allocation rate for LS-MCPP. Semantic corrections update the shared planning state without requiring repeated transmission of the complete map.
\end{abstract}


\section{Introduction}

Disaster response operations increasingly utilize heterogeneous teams of Unmanned Aerial Vehicles (UAVs) and ground robots \cite{jsan13060081,ground_AIR1:zhang2025airgroundcollaborativerobotsrescue} to explore environments that are unsafe or inaccessible to human responders. UAVs are more effective for surveying flooded or structurally collapsed areas, while ground robots provide close-range inspection where terrain conditions permit, including regions with loose debris, dense vegetation, or obstructing tree branches. However, these platforms can struggle to function as a coherent team. Aerial robots may observe regions that a ground robot cannot safely traverse due to structural collapse, unstable terrain, or flooding, leading to spatial mismatches, infeasible task assignments, and redundant exploration. These challenges highlight a critical disconnect between conventional Coverage Path Planning (CPP), which typically assumes homogeneous mobility and centralized global maps, and emerging search-and-rescue (SAR) deployments in which heterogeneous robots coordinate through distributed, edge-centric architectures envisioned for 6G-enabled systems.

Effective collaboration therefore depends not only on robot autonomy but also on exchanging mission information needed for coordination. Research on 6G networks envisions communication, edge computing, and AI inference as integrated components of distributed decision loops rather than as isolated functions \cite{EU_6G:1bf3fab0bd4540e3b2d2e8f969f63c1a,semantic_edge_6g:Zhang_2025}. For SAR robot teams, the value of an observation lies not in accurately reconstructing the original sensor data, but in whether it changes the semantic state used for terrain traversability and coverage assignment. The proposed mechanism therefore integrates AI-derived semantics across the entire coordination loop: a deep learning model extracts terrain semantics, these AI-derived constraints dictate capability-aware path planning, and the resulting state changes are exchanged through application-level goal-oriented communication rather than continuous imagery \cite{semantic_6g_survey_10849550}.

Communication capabilities alone are insufficient if the underlying planner does not account for heterogeneous traversability \cite{survey_com_robot:gielis2022criticalreviewcommunicationsmultirobot, bravoarrabal2025strengtheningmultirobotsystemssar}. Disaster environments often contain adjacent regions with different traversability constraints, including damaged infrastructure, debris, and flooding, requiring platform-specific decisions about which robots can safely and effectively access each area \cite{rescurenet:rahnemoonfar2023rescuenet, floodnet:rahnemoonfar2020floodnet}. Coverage planning must therefore treat reachability as a primary constraint rather than rely solely on efficiency or spatial proximity. Existing research has largely considered semantic perception \cite{db_cam_pos1:sirma2025drespnetuavdatasetyolov8drn,db_cam_pos2:le20253dsemanticsegmentationpostdisaster}, multi-robot coverage planning \cite{survey_mcpp1:GALCERAN20131258,survey_mcpp2:almadhoun-2019}, and communication efficiency \cite{survey_com_robot:gielis2022criticalreviewcommunicationsmultirobot,5g_6g1:ghassemian20266gempoweringfuturerobotics} as separate problems.  However, simply combining these components is insufficient because standard planners typically assume a shared terrain graph and do not enforce robot-specific traversability constraints, potentially resulting in infeasible path assignments. A gap therefore remains between identifying terrain semantics, translating them into robot-specific coverage assignments, and efficiently exchanging the resulting state during distributed operation.

\begin{figure*}[htb]
    \centering
    \includegraphics[width=1.0\textwidth]{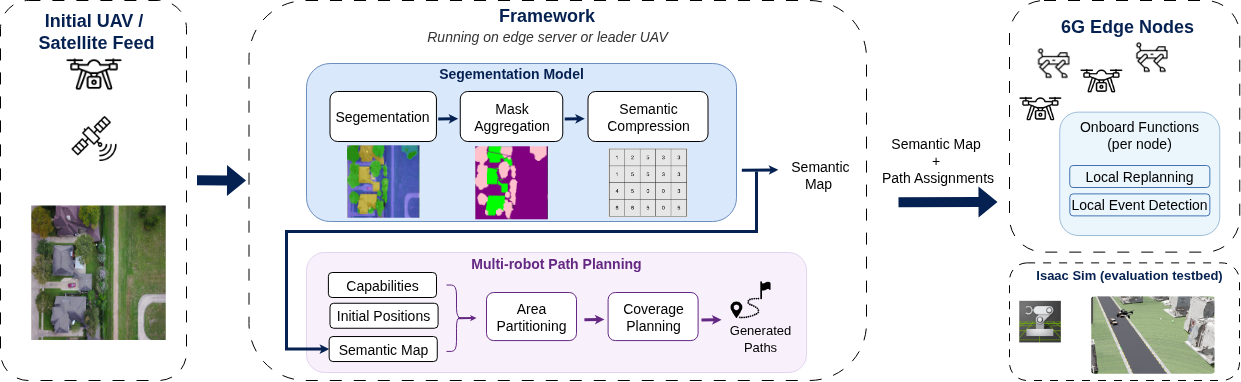}
    \caption{Framework overview. Aerial or satellite imagery is converted into a semantic map and combined with robot capabilities and initial positions for area partitioning and coverage planning. The semantic map and path assignments are distributed to aerial and ground robots, where onboard functions support local event detection and replanning. Isaac Sim provides the evaluation testbed.}
    \vspace{-15pt}
    \label{fig:overview}
\end{figure*}

To address these challenges, this paper makes the following contributions: (i) an edge-centric coordination framework that structurally couples semantic perception, capability-aware coverage planning, and semantic information exchange for heterogeneous SAR robot teams, as shown in Figure~\ref{fig:overview}; (ii) a reachability-constrained Voronoi decomposition and coverage-planning method that embeds platform-specific traversability constraints directly into the geodesic expansion to ensure feasible coverage and support local replanning; and (iii) a goal-oriented semantic state-sharing mechanism that exchanges incremental semantic-map corrections rather than continuous imagery.

\section{Related Work}

\subsection{Perception and Semantic Mapping for Disaster Robotics}
Semantic perception from overhead imagery is increasingly used in search-and-rescue (SAR) operations \cite{db_cam_pos1:sirma2025drespnetuavdatasetyolov8drn,Semantics-basedSituationalAwareness:ruan2025frameworksemanticsbasedsituationalawareness}. Disaster datasets such as RescueNet \cite{rescurenet:rahnemoonfar2023rescuenet} and FloodNet \cite{floodnet:rahnemoonfar2020floodnet} have advanced the capability of deep learning models to segment structural damage, debris, and flooded areas. However, existing work generally uses these semantic labels for damage assessment rather than translating them into robot-specific traversability constraints for coverage planning. Large-scale satellite datasets such as xBD extend assessment to regional building damage, yet their building-focused annotations do not describe the local access conditions required for ground navigation \cite{xbd_xview2:gupta2019xbddatasetassessingbuilding}. In this context, there remains a need for semantic representations that translate overhead scene information into robot-specific traversability constraints for coverage planning.

\subsection{Coverage and Path Planning for Heterogeneous Multi-Robot Systems}
Coverage Path Planning (CPP) aims to achieve systematic exploration of an environment while minimizing redundant motion and execution time. Classical methods primarily address single robots \cite{CPP1:10.1007/978-1-4471-1273-0_32} or homogeneous teams \cite{CCP2_deep:9724591,CPP_DARP:kapoutsis_2017_darp} and typically use a common motion model over geometrically defined free space. Multi-robot CPP (mCPP) extensions scale coverage through task allocation and area decomposition \cite{CPP3:s24237482}, with DARP and Voronoi partitioning balancing workload across disjoint regions \cite{CPP_DARP:kapoutsis_2017_darp,Voronoi-BasedSpacePartitioning}. LS-MCPP applies local search to reduce coverage makespan, while MFC constructs rooted forest covers over weighted or unweighted terrain \cite{tang2024lsmcpp,algo_mfc:5629446}. Both operate on a shared terrain graph and do not encode robot-specific semantic traversability during allocation. Heterogeneous systems have been addressed through role-specific UAV-UGV collaboration in partially known environments \cite{HeterogeneousMulti-RobotCollaboration:machines12030200} and graph coverage with explicit movement and proximity constraints \cite{mutzari2025heterogeneous}. The former coordinates platform roles around predefined coverage paths, whereas the latter optimizes constrained covering tours on an input graph. However, neither translates perception-derived terrain labels into robot-specific reachability constraints. Our framework addresses this gap by embedding platform-specific traversability directly into semantic-grid decomposition.

\subsection{AI-Native 6G Semantic Coordination}

Reliable, low-latency communication is a critical enabler for multi-robot SAR operations, particularly in time-critical and safety-sensitive missions. Research toward sixth-generation (6G) wireless networks envisions communication infrastructures that support edge intelligence, sub-millisecond radio latency, and ultra-reliable connectivity \cite{EU_6G:1bf3fab0bd4540e3b2d2e8f969f63c1a,5g_6g1:ghassemian20266gempoweringfuturerobotics}. Within this vision, the impracticality of continuous raw sensor streaming for large heterogeneous teams motivates Semantic Communication (SemCom), which shifts the objective from reconstructing source data to delivering task-relevant information for downstream decisions \cite{semantic_6g_survey_10849550}. Such goal-oriented communication is especially relevant in dynamic disaster environments, where delayed semantic updates can result in inconsistent semantic maps, redundant coverage, or infeasible assignments.

Edge intelligence provides a practical basis for implementing semantic coordination in multi-robot SAR. Edge-enabled architectures allow perception and planning to remain onboard each robot, while nearby computing services can maintain and distribute the shared environmental state \cite{semantic_edge_6g:Zhang_2025}. This arrangement aligns with the AI-native 6G vision, in which communication and edge computing support distributed AI-based perception and planning. In the proposed framework, incremental semantic-map corrections reduce the need to transmit continuous imagery while updating the traversability information used for coverage assignment and replanning.

\section{Proposed Approach}
To address capability-infeasible allocations in heterogeneous teams, the proposed framework converts aerial observations into semantic grid maps for reachability-constrained decomposition, local path planning, and incremental state sharing.

\subsection{Semantic Environment Representation}
\begin{figure*}
    \centering
    \includegraphics[width=1\linewidth]{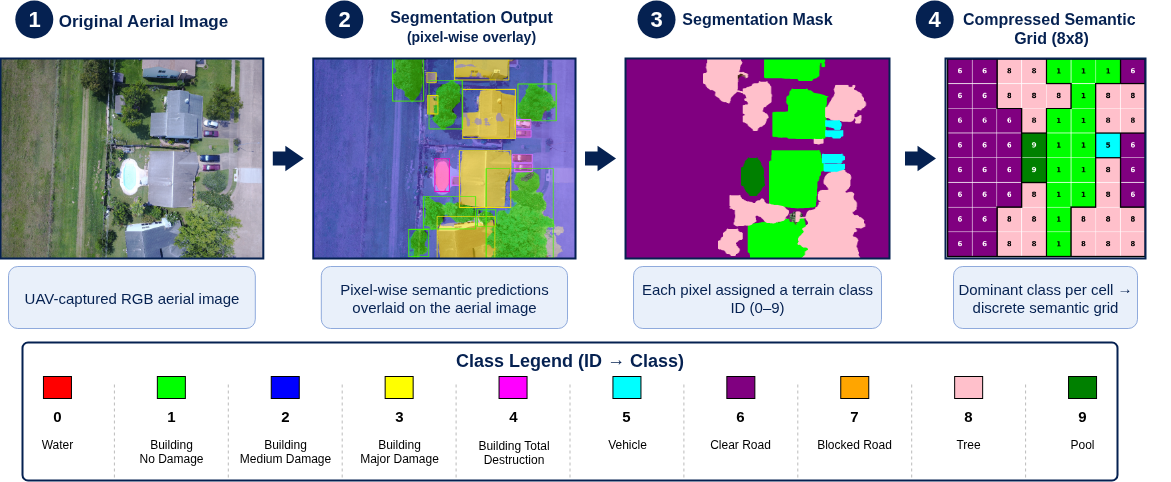}
    \caption{Semantic perception pipeline from aerial imagery to a grid-based semantic map. Pixel masks are aggregated by dominant class into a discrete semantic grid for capability-aware multi-robot planning.}
    \label{fig:segmentation_flow}
\end{figure*}

The disaster environment is represented as a semantic grid map derived from aerial imagery using a YOLOv11 segmentation model \cite{yolo11_ultralytics}, which produces pixel masks for terrain and structural elements relevant to search-and-rescue operations. To support coverage planning, the segmented output is partitioned into an operator-defined $R \times C$ grid, whose dimensions control the spatial resolution of the map. Each grid cell is assigned the most frequent predicted semantic class among its labeled pixels, producing a compact representation of obstacles and terrain semantics. Cells without a predicted terrain label are marked as unknown, treated as impassable during planning, and excluded from the coverage denominator. The resulting grid serves as the input for capability-aware area decomposition and path planning, as shown in Figure~\ref{fig:segmentation_flow}.

\subsection{Reachability-Constrained Area Decomposition and Path Planning}
Area decomposition and path planning are performed over the semantic map to assign disjoint coverage regions and generate feasible paths for each agent. The environment is first partitioned into robot-specific regions based on terrain information from the semantic map and each robot’s traversability constraints. Local coverage targets are then selected independently within each assigned region.

\subsubsection{Reachability-Constrained Decomposition}
Let $G \in \mathbb{Z}^{R \times C}$ denote a discrete semantic map, where each cell encodes a terrain class inferred from aerial semantic segmentation. Each robot $r_i$ is associated with a capability set 
\begin{equation}
\mathcal{T}_i = \{t_1, t_2, \dots\},
\end{equation} representing the semantic classes admitted by the platform's mobility constraints and the operator-defined requirements of the current mission.

The environment is partitioned into disjoint, robot-specific coverage regions using a discrete geodesic (shortest-path) Voronoi decomposition \cite{GEODESICPROBLEM:mitchell1987discrete} computed on the semantic grid. Geodesic distances are computed along the grid graph while respecting robot-specific terrain traversability, thereby accounting for obstacles and heterogeneous capabilities. For each robot $r_i$, a wavefront expansion is performed using Breadth-First Search (BFS), initialized from the robot’s start cell. During expansion, neighboring cells are explored only if they satisfy the constraint $G(x,y) \in \mathcal{T}_i$. Cells that violate the robot’s capability set are treated as impassable, embedding terrain-dependent traversability directly into the distance computation. This process yields a distance map $d_i(x,y)$ representing the minimum number of grid steps required for robot $r_i$ to reach cell $(x,y)$. Each semantic cell that is reachable by at least one robot is assigned to the robot with the minimum geodesic cost according to
\begin{equation}
\pi(x,y) = \arg\min_{i:d_i(x,y)<\infty} \; d_i(x,y).
\end{equation}
Cells that cannot be reached from any robot's starting position through terrain permitted by its capability set remain unassigned.

\subsubsection{Local Coverage Path Planning}
Following the area decomposition, an initial coverage path is computed for each robot to cover its assigned region. The robot executes this path locally using the semantic grid map. During execution, onboard observations update the semantic map, and the path is replanned when changes in terrain or obstacles are detected. Path construction proceeds incrementally by selecting the nearest uncovered assigned cell in Manhattan distance, with feasible routes generated using capability-constrained A* search over the semantic grid. Uncovered assigned cells encountered along each path segment are marked as covered, enabling implicit path stitching that reduces backtracking and traversal overhead while covering the reachable cells assigned to each robot.

\subsection{Goal-Oriented Semantic State Sharing}
Semantic information exchange is performed through compact, grid-based updates rather than raw imagery or complete semantic-map broadcasts, reflecting the bandwidth-efficient and latency-sensitive communication patterns envisioned for emerging 6G-enabled collaborative robotic systems. Each ground robot and UAV maintains a local semantic grid, initialized from a shared map and updated through onboard perception. When terrain traversability changes or previously unknown obstacles are observed, only the affected semantic cells are shared. These updates encode the location and revised semantic state of the affected cells for integration into each agent’s local map. Due to the distributed architecture, agents may temporarily operate with incomplete or outdated semantic maps. Each robot therefore plans from its local representation and invokes onboard replanning when newly observed terrain conflicts with the current map. Evaluating robustness to communication delays and inter-agent map inconsistencies is left for future work. Overall, lightweight semantic-cell updates combine local perception and replanning in an application-level coordination loop motivated by the AI-native 6G vision.

\section{Datasets}
Robust semantic understanding of disaster environments is essential for reliable multi-robot coverage planning. However, labeled data for post-disaster scenarios remain scarce and inconsistent across datasets. To address this limitation, a custom dataset was constructed by integrating three open-source datasets that capture complementary viewpoints, scales, and disaster types. The final dataset comprises a total of 5,234 images partitioned into training, validation, and testing sets using an 80/10/10 ratio. Combining these sources provides varied terrains, viewpoints, spatial scales, and disaster conditions relevant to heterogeneous robot teams. RescueNet provides high-resolution post-Hurricane Michael UAV imagery with dense annotations of structural damage and navigable ground surfaces \cite{rescurenet:rahnemoonfar2023rescuenet}. The Semantic Segmentation Satellite Imagery Dataset contains 261 high-resolution urban images from Houston, Texas \cite{semantic_segmentation_satellite_imagery:alchimowicz_2022} and was included to provide additional examples of intact buildings and urban structures. FloodNet contains approximately 2,343 images collected after Hurricane Harvey, including annotations for flooded roads and buildings \cite{floodnet:rahnemoonfar2020floodnet}.

\subsection{Data Pre-processing}
All images and annotation masks were converted to a common format and resized to a uniform resolution compatible with the YOLO-based segmentation architecture. The source datasets use partially overlapping labels at different levels of granularity. We therefore mapped their annotations to a common taxonomy of 10 terrain classes (Table~\ref{tab:terrain_classes}). The taxonomy retains distinctions in structural damage and terrain accessibility needed for robot-specific traversability reasoning and coverage planning.

\begin{table}[t]
\centering
\caption{Terrain Classes Used for Semantic Segmentation}
\label{tab:terrain_classes}
\setlength{\tabcolsep}{3pt}
\renewcommand{\arraystretch}{1.15}
\small
\begin{tabularx}{\columnwidth}{@{}c p{3.5cm} Y@{}}
\toprule
\textbf{ID} & \textbf{Class} & \textbf{Description} \\
\midrule
0 & Water & Flooded areas, standing water, and ponds \\
1 & Building No Damage & Intact structures with no visible damage \\
2 & Building Medium Damage & Visible structural or roof damage; structure remains standing \\
3 & Building Major Damage & Partial collapse or severe structural failure \\
4 & Building Total Destruction & Rubble or fully collapsed structures \\
5 & Vehicle & Cars and trucks \\
6 & Clear Road & Unobstructed roads and lands \\
7 & Blocked Road & Roads obstructed by debris \\
8 & Tree & Vegetation and trees \\
9 & Pool & Man-made swimming pools \\
\bottomrule
\end{tabularx}
\end{table}

\section{Experimentation}
\subsection{Segmentation Model}
The semantic segmentation model was trained on a composite dataset designed to capture heterogeneous spatial scales and viewpoints typical of aerial disaster assessment. Performance was evaluated using pixel-level precision, recall, and micro F1-score. Evaluation on $523$ held-out test images yielded a precision of $0.883$, a recall of $0.837$, and a micro F1-score of $0.859$. The higher precision indicates fewer false-positive terrain labels, while dominant-class aggregation may limit the influence of isolated pixel-level errors. Lower performance was observed for sparsely represented classes, including Building Medium Damage, Building Major Damage, Building Total Destruction, Blocked Road, and Pool, likely reflecting limited training samples and visual ambiguity. Building damage classes also exhibited lower accuracy under oblique viewpoints, consistent with the dominance of top-down imagery in the training data \cite{db_cam_pos1:sirma2025drespnetuavdatasetyolov8drn,db_cam_pos2:le20253dsemanticsegmentationpostdisaster}. However, misclassification of a cell's dominant terrain class can still affect traversability decisions and subsequent coverage assignments.

\subsection{Isaac Sim Feasibility Evaluation}
\begin{figure}[htp]
    \centering
    \includegraphics[width=1\linewidth]{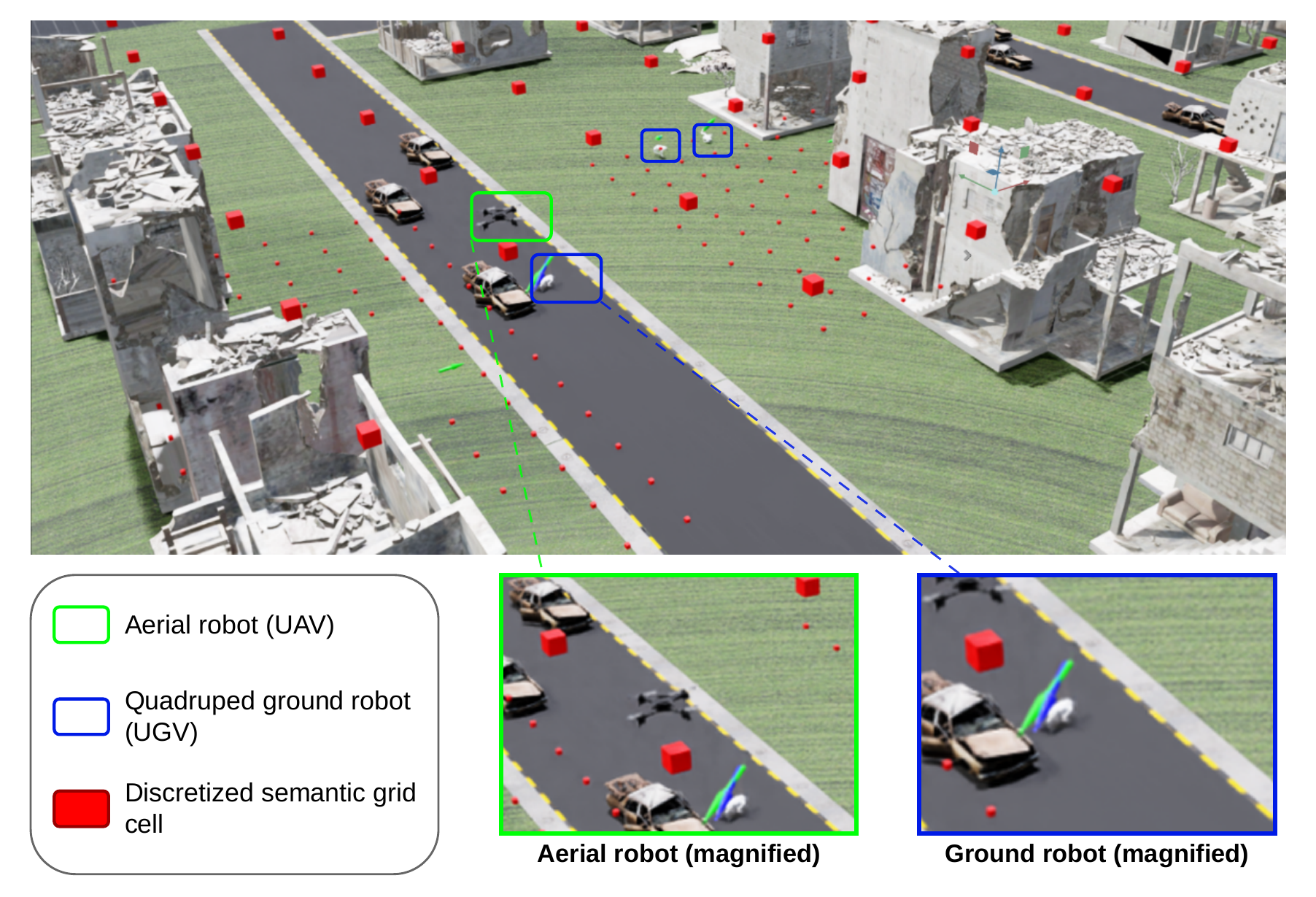}
    \caption{Multi-robot semantic coverage in Isaac Sim. Snapshot of a simulated disaster environment showing aerial robots (green) and quadruped ground robots (blue). Red cubes denote discretized semantic grid cells used to encode robot-specific traversability and coverage maps. Magnified insets show one aerial and one ground robot during execution.}
    \label{fig:sim_3dones1}
\end{figure}

A representative post-disaster urban environment was constructed in Isaac Sim \cite{NVIDIA_Isaac_Sim} for a controlled feasibility evaluation of the proposed semantic perception pipeline and heterogeneous coverage-planning algorithm. The simulated scene emulates search-and-rescue conditions, including partially collapsed buildings and debris-obstructed roads. Figure~\ref{fig:sim_3dones1} shows the aerial and ground robot deployment together with the discretized semantic grid representation.

Ten Isaac Sim trials were conducted using heterogeneous teams of quadruped ground robots and aerial robots as a closed-loop feasibility evaluation. Semantic observations were converted into grid maps and used for capability-aware assignment and path planning. The initial planning pass assigned approximately $85\%$ of grid cells, with the remaining cells left unassigned because of conservative capability constraints, segmentation uncertainty, and variations in robot starting positions. During execution, local traversal conflicts were identified in fewer than $10\%$ of assigned cells, primarily where dominant-class aggregation concealed non-traversable content within cells whose dominant label was feasible for the assigned robot. These conflicts were subsequently corrected through semantic-map updates and local replanning. The larger semantic-grid benchmark in Section~\ref{subsec:baseline} provides the reproducible quantitative comparison against other planners.

\subsection{Quantitative Planner Benchmark}
\label{subsec:baseline}
We evaluate the proposed method against two baseline groups: classical capability-blind decomposition and graph-based mCPP planners. The classical baselines are DARP \cite{CPP_DARP:kapoutsis_2017_darp} and a modified form of boustrophedon coverage \cite{CPP1:10.1007/978-1-4471-1273-0_32}, while the graph-based planners are LS-MCPP \cite{tang2024lsmcpp} and Multi-Robot Forest Coverage (MFC) \cite{algo_mfc:5629446}. The modified baseline divides the grid into equal contiguous column regions without using terrain capabilities, then orders each robot's feasible targets using an alternating row-wise sweep. The benchmark uses one ground robot and one UAV with $\mathcal{T}_{\mathrm{ground}}=\{6,7,8\}$ and $\mathcal{T}_{\mathrm{UAV}}=\{0,1,2,3,4,5,6,7,9\}$, using the class identifiers in Table~\ref{tab:terrain_classes}. Each capability set combines platform mobility constraints, which determine where the robot can physically move, with operator-defined mission constraints. In the proposed method, these sets constrain both decomposition and path construction. Cells outside a robot's set are neither traversed nor overflown nor assigned for coverage. The ground-robot set describes admissible terrain traversal, while the UAV set describes admissible aerial movement and coverage. These sets are scenario-specific and can be adapted to other platforms or missions. All planners are evaluated on identical $20\times20$ semantic grids across a common set of $814$ randomized configurations. Of $900$ attempted configurations, $85$ were excluded because the DARP implementation either failed on disconnected free-space grids or did not converge, and one configuration was incompatible with MFC's contracted-root representation. Coverage is the percentage of segmented cells covered by a capable robot. Capability-infeasible allocation is the percentage of robot--cell allocation pairs outside the corresponding capability set; for LS-MCPP and MFC, allocations are recovered from the unique grid cells in each completed coverage walk.

\begin{table}[t]
\centering
\caption{Mean planner performance over the common configuration set.}
\label{tab:baseline_many}
\scriptsize
\setlength{\tabcolsep}{4pt}
\begin{tabular}{@{}lcc@{}}
\toprule
Planner & Cov.\,(\%)$\uparrow$ & Infeas. alloc.\,(\%)$\downarrow$ \\
\midrule
Proposed & $\mathbf{91.5}$ & $\mathbf{0.0}$ \\
DARP \cite{CPP_DARP:kapoutsis_2017_darp} & $74.2$ & $21.0$ \\
Modified boustrophedon \cite{CPP1:10.1007/978-1-4471-1273-0_32} & $71.9$ & $22.6$ \\
LS-MCPP \cite{tang2024lsmcpp} & $78.8$ & $21.5$ \\
MFC rooted-tree cover \cite{algo_mfc:5629446} & $79.1$ & $21.5$ \\
\bottomrule
\end{tabular}
\end{table}

Table~\ref{tab:baseline_many} reveals two consistent patterns. First, DARP and the modified boustrophedon baseline produce capability-infeasible allocation rates of $21.0\%$ and $22.6\%$, respectively, while LS-MCPP and MFC both produce rates of approximately $21.5\%$. The proposed capability-aware method produces none. Second, these infeasible allocations are accompanied by lower feasible coverage: $74.2\%$ for DARP, $71.9\%$ for the modified boustrophedon baseline, $78.8\%$ for LS-MCPP, and $79.1\%$ for MFC, compared with $91.5\%$ for the proposed method. Although the graph-based planners improve coverage over the classical partitions, their common traversability model does not prevent platform-infeasible allocations. By incorporating robot-specific geodesic reachability during decomposition, the proposed method assigns cells only to robots with a capability-valid path from their starting position.

\subsection{Discretization Resolution Analysis}
\label{subsec:resolution}

\begin{figure}[t]
\centering
\includegraphics[width=\linewidth]{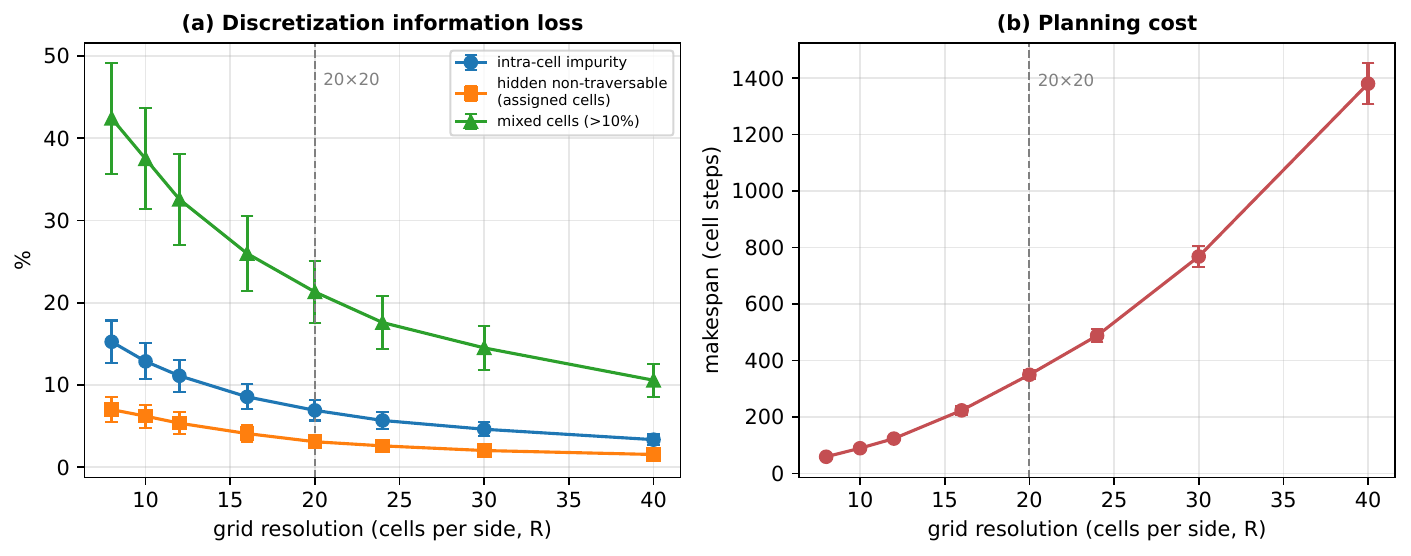}
\caption{Effect of grid resolution on discretization fidelity and grid-step makespan across 80 scenes. (a) Finer grids reduce intra-cell class mixing, hidden non-traversable area, and the proportion of highly mixed cells. (b) Planning makespan increases rapidly with resolution. }
\label{fig:e3_resolution}
\end{figure}

The semantic grid compresses each cell to its dominant class, so mixed cells can hide small non-traversable regions. To quantify this information loss, we evaluate grid resolution from $8\times8$ to $40\times40$ over $80$ scenes. Intra-cell impurity measures the proportion of pixels that differ from a cell's dominant semantic class, while the hidden non-traversable fraction measures non-traversable pixels concealed by a traversable dominant class. Intra-cell impurity decreases from $15.2\%$ at $8\times8$ to $3.3\%$ at $40\times40$, while the hidden non-traversable fraction falls from $7.0\%$ to $1.5\%$. The simulation experiments use a $20\times20$ grid, for which the corresponding impurity and hidden non-traversable fractions are $6.9\%$ and $3.1\%$, respectively. Finer grids also increase the number of discrete path steps, with the maximum per-robot path length (makespan) growing from roughly $59$ to $1{,}380$ grid-cell steps across the resolution sweep. The appropriate resolution is not fixed, as it scales with the covered area and altitude. Images captured at a higher altitude represent a larger physical area, so the grid dimensions should be increased to keep the area represented by each cell approximately consistent. Figure~\ref{fig:e3_resolution} summarizes the resolution sweep.

\subsection{Application-Payload Analysis for Semantic State Sharing}
We compare the application payload required for uncompressed aerial-image streaming, periodic and event-triggered full semantic-map sharing, and delta-based semantic updates. Raw image transmission is treated as a reference upper bound on communication cost. An RGB image of resolution $1920 \times 1080$ requires approximately $6.2$~MB per frame, resulting in a data rate of $62$~MB/s per robot at $10$~Hz, excluding compression and protocol overhead. The aggregate source traffic increases linearly with the number of robots. As lower-complexity baselines, we consider sharing complete semantic grid maps. For a $20 \times 20$ grid with one byte per cell, a full update requires $400$~bytes, corresponding to $4.0$~kB/s at the same frequency. Full-map sharing therefore reduces the application payload by approximately four orders of magnitude relative to uncompressed image streaming, but repeatedly transmits unchanged cells. To isolate the effect of delta encoding from that of event-triggered transmission, we additionally consider an event-triggered full-map baseline, in which the complete semantic map is transmitted whenever a semantic change is detected.

The proposed exchange mechanism transmits an update only when onboard observation identifies a difference from the stored semantic map. Let $f$ denote the semantic observation frequency, $q$ the probability that an observation produces an update, $\bar{n}$ the mean number of changed cells in a non-empty update, $b$ the bytes used to encode each changed cell, and $h$ the application-message header. The expected application-payload rate is
\begin{equation}
R_{\Delta}=fq(\bar{n}b+h).
\end{equation} 
Using $f=10$~Hz, $b=3$~B for a two-byte cell index and one-byte semantic label, $h=4$~B, and $\bar{n}=1$, the payload ranges from $0$ to $70$~B/s depending on the frequency of semantic corrections. Under the high-update assumption $q=0.70$, event-triggered full-map sharing requires approximately $2.8$~kB/s, while the proposed delta updates require $49$~B/s. This corresponds to a payload reduction of approximately $57\times$ relative to event-triggered full-map sharing and $82\times$ relative to periodic full-map sharing.

Each semantic correction updates the terrain state used for traversability reasoning and coverage assignment. The resulting exchange is therefore goal-oriented at the application layer, as robots share planning-relevant changes rather than continuous imagery. These values quantify application payload under the stated encoding and update assumptions, rather than end-to-end network performance.

\section{Conclusion}
This paper presented an edge-centric semantic coverage framework for heterogeneous SAR robot teams, in which terrain labels derived from aerial imagery constrain robot-specific reachability before spatial decomposition and incremental corrections update the shared planning state. The segmentation model achieved a micro F1-score of $0.859$ on $523$ held-out images. In the common benchmark, the proposed method achieved $91.5\%$ coverage with no capability-infeasible allocations, compared with $78.8\%$ coverage and a $21.5\%$ capability-infeasible allocation rate for LS-MCPP. Representing the resulting planning state as a $20\times20$ semantic grid reduced the application payload by approximately four orders of magnitude relative to uncompressed imagery. Under the high-update assumptions used in the payload analysis, the delta-update model yields $49$~B/s, a reduction factor of approximately $82\times$ relative to periodic full-map sharing. These results show that semantic corrections are exchanged for their effect on traversability reasoning, coverage assignment, and replanning rather than to reconstruct the underlying imagery. Coupled with onboard inference and edge-centric state sharing, this decision loop provides an application-level example of AI-enabled goal-oriented communication aligned with the emerging AI-native 6G vision. Future evaluation will examine the complete loop on physical heterogeneous platforms under communication delay and inter-agent map inconsistency.

\section*{Data and Code Availability}
Due to licensing constraints on the source datasets, we do not redistribute the merged dataset or its annotations. The label-mapping and preprocessing scripts used to reproduce the merged dataset from the original sources, together with instructions for obtaining those sources, are available at \url{https://github.com/adhafer/semcap-cpp}.

\bibliographystyle{IEEEtran}
\bibliography{references}
\end{document}